\documentclass[letterpaper, 10 pt, conference,dvipsnames]{ieeeconf}  

\IEEEoverridecommandlockouts                              
\usepackage[utf8]{inputenc}
\usepackage[T1]{fontenc}

\usepackage{color}
\usepackage{xcolor}
\usepackage{commath}
\usepackage{booktabs}

\usepackage{graphicx} 
\usepackage{amsmath} 
\usepackage{amssymb}  
\usepackage{cancel}
\usepackage{verbatim}
\usepackage{algorithm}
\usepackage{algpseudocode}
\usepackage{float}
\usepackage{tikz}
\usetikzlibrary{automata,positioning}
\tikzstyle{node}=[align=center]

\usepackage{tabularx}
\usepackage{diagbox} 
\usepackage{cancel}
\usepackage{bm}
\renewcommand{\arraystretch}{1.5}

\usepackage{mathtools}

\usepackage{comment}
\usepackage{subcaption}
\usepackage{hyperref}
\usepackage{siunitx}
\usepackage{soul}
\usepackage{svg}

\usetikzlibrary{automata,positioning}
\tikzstyle{node}=[align=center]

\usepackage[precision=2, unit=mm]{lengthconvert}

\let\labelindent\relax
\usepackage[inline]{enumitem}

\usepackage{amsmath}

\newcommand{\R}{{\mathbb{R}}}
\newcommand{\bx}{{\bm{x}}}
\newcommand{\bu}{{\bm{u}}}
\newcommand{\bp}{{\bm{p}}}
\newcommand{\be}{{\bm{e}}}
\newcommand{\bt}{{\bm{t}}}
\newcommand{\bz}{{\bm{z}}}
\newcommand{\bq}{{\bm{q}}}

\newcommand{\bR}{{\mathbf{R}}}
\newcommand{\bJ}{{\mathbf{J}}}
\newcommand{\Log}{{\operatorname{Log}}}
\newcommand{\Exp}{{\operatorname{Exp}}}
\newcommand{\RBF}{{\operatorname{RBF}}}

\usepackage{acro}
\usepackage{xspace}

\DeclareAcronym{IK}{
    short = IK,
    long = Inverse Kinematics
}

\DeclareAcronym{CBF}{
    short = CBF,
    long = Control Barrier Function
}
\newcommand{\cbfs}{\ac{CBFs}\xspace}
\DeclareAcronym{CBFs}{
    short =  CBFs,
    long = Control Barrier Functions
}

\newcommand{\apf}{\ac{APF}\xspace}
\DeclareAcronym{APF}{
    short = APF,
    long = Artificial Potential Field
}

\newcommand{\rbf}{\ac{RBF}\xspace}
\DeclareAcronym{RBF}{
    short = RBF,
    long = Relaxed Barrier Function
}

\DeclareAcronym{RMSE}{
    short = RMSE,
    long = Root Mean Square Error
}

\DeclareAcronym{std}{
    short = std,
    long = standard deviation
}

\DeclareAcronym{wrt}{
    short = w.r.t.,
    long = with respect to
}

\DeclareAcronym{ctt}{
    short = CTT,
    long = Cartesian Trajectory Tracking
}

\newcommand{\nn}{\ac{NN}\xspace}

\DeclareAcronym{NN}{
    short = NN,
    long = Neural Network
}
\DeclareAcronym{FC}{
    short = FC,
    long = Fully Connected
}

\DeclareAcronym{CNN}{
    short = CNN,
    long = Convolutional \acl{NN}
}

\DeclareAcronym{MLP}{
    short =MLP,
    long = Multi-Layer Perceptron
}
\DeclareAcronym{BCE}{
    short = BCE,
    long = Binary Cross Entropy
}
\DeclareAcronym{RL}{
    short = RL,
    long = Reinforcement Learning
}
\DeclareAcronym{IL}{
    short = IL,
    long = Imitation Learning
}

\DeclareAcronym{DOF}{
    short = DoF,
    long = Degree of Freedom,
    long-plural-form = Degrees of Freedom
}

\newcommand{\mpc}{\ac{MPC}\xspace}
\DeclareAcronym{MPC}{
    short = MPC,
    long = Model Predictive Control
}
\newcommand{\ttmpc}{\ac{TT-MPC}\xspace}
\DeclareAcronym{TT-MPC}{
    short = TT-\acs{MPC},
    long = Trajectory Tracking \acl{MPC}
}
\newcommand{\mpcc}{\ac{MPCC}\xspace}
\DeclareAcronym{MPCC}{
    short = MPCC,
    long = Model Predictive Contouring Control
}
\newcommand{\rmpcc}{\ac{RMPCC}\xspace}
\DeclareAcronym{RMPCC}{
    short = R\acs{MPCC},
    long = Reactive \acl{MPCC}
}

\DeclareAcronym{MPFC}{
    short = MPFC,
    long = Model Predictive path-Following Control
}

\DeclareAcronym{NMPC}{
    short = N-\acs{MPC},
    long = Nonlinear \acl{MPC}
}
\newcommand{\qp}{\ac{QP}\xspace}

\DeclareAcronym{QP}{
    short = QP,
    long = Quadratic Programming
}

\newcommand{\sqp}{\ac{SQP}\xspace}
\DeclareAcronym{SQP}{
    short = SQP,
    long = Sequential \acl{QP}
}
\newcommand{\ocp}{\ac{OCP}\xspace}
\DeclareAcronym{OCP}{
    short = OCP,
    long = Optimal Control Problem
}

\DeclareAcronym{RTI}{
    short = RTI,
    long = Real-Time Iteration
}
\DeclareAcronym{LQR}{
    short = LQR,
    long = Linear-Quadratic Regulator
}

\DeclareAcronym{NLP}{
    short = NLP,
    long = NonLinear Programming
}

\DeclareAcronym{KKT}{
    short = KKT,
    long = Karush-Kuhn-Tucker
}
\DeclareAcronym{RGBD}{
    short = RGBD,
    long = RGB + Depth
}

\DeclareAcronym{COM}{
    short = CoM,
    long = Center of Mass
}

\DeclareAcronym{CPU}{
    short = CPU,
    long = Central Processing Unit
}
\DeclareAcronym{GPU}{
    short = GPU,
    long = Graphical Processing Unit
}
\include{h2t_def}

\title{\LARGE \bf
Reactive Model Predictive Contouring Control for Robot Manipulators
}

\author{Junheon Yoon$^{1}$, Woo-Jeong Baek$^{2}$, and Jaeheung Park$^{1,3}$
\thanks{*This work was supported by the Technology Innovation Program funded by the Ministry of Trade, Industry \& Energy (MOTIE) (RS-2024-00423940), the Institute of Information \& Communications Technology Planning \& Evaluation (IITP) grant funded by the Korea government (MSIT) (No.RS-2024-00459435), and the Imdang Scholarship \& Cultural Foundation.}
\thanks{$^{1}$Department of Intelligence and Information, Graduate
School of Convergence Science and Technology, Seoul National University, Seoul 08826, Republic of Korea. {\tt\small yoonjh98@snu.ac.kr}} 
\thanks{$^{2}$Artificial Intelligence Institute(AIIS), Seoul National University, Republic of Korea and Institute for Anthropomatics and Robotics (IAR-IPR), Karlsruhe Institute of Technology (KIT), Germany.  {\tt\small wjbaek@snu.ac.kr}}
\thanks{$^{3}$Department of Intelligence and Information, Graduate School of Convergence Science and Technology, ASRI, AIIS, Seoul National University, Seoul 08826, Republic of Korea, and the Advanced Institutes of Convergence Technology (AICT) Suwon 16229, Republic of Korea. {\tt\small park73@snu.ac.kr}}
\thanks{The authors like to thank Jaehyun Kim for his guidance and fruitful discussions throughout this work.}%
}

\begin{document}
\graphicspath{ {./img/} }

\maketitle
\thispagestyle{empty}
\pagestyle{empty}
\begin{abstract}
This contribution presents a robot path-following framework via Reactive Model Predictive Contouring Control (RMPCC) that successfully avoids obstacles, singularities and self-collisions in dynamic environments at 100 Hz. Many path-following methods rely on the time parametrization, but struggle to handle collision and singularity avoidance while adhering kinematic limits or other constraints. Specifically, the error between the desired path and the actual position can become large when executing evasive maneuvers. Thus, this paper derives a method that parametrizes the reference path by a path parameter and performs the optimization via RMPCC. In particular, Control Barrier Functions (CBFs) are introduced to avoid collisions and singularities in dynamic environments. A Jacobian-based linearization and Gauss-Newton Hessian approximation enable solving the nonlinear RMPCC problem at 100 Hz, outperforming state-of-the-art methods by a factor of 10. Experiments confirm that the framework handles dynamic obstacles in real-world settings with low contouring error and low robot acceleration.
\end{abstract}

\section{INTRODUCTION}
Robotic manipulators are widely used in various industrial and research applications, including welding \cite{my2019inverse}, painting \cite{goller2022model}, and peg-in-hole assembly \cite{lee2021search}, which demand precise \emph{path-following}. Unlike point-to-point tasks, path-following requires \emph{continuous adherence} to an entire trajectory, increasing constraints. 
Accurate path-following is required for safe and efficient task performance, and constraints such as kinematic limitations and safety factors including singularity and collision avoidance must be addressed. One possibility to address these problems is to parameterize the reference path over time and track it, respectively (\cite{lee2023real, zhu2023real}). This approach is usually organized into two hierarchical controllers:  A high-level trajectory generator creates a collision-free, singularity-free trajectory by considering robot dynamics at a lower frequency, and a low-level trajectory tracking controller computes control inputs at a higher frequency to track it. However, such an approach is often complex to design. In addition, addressing collision avoidance in the high-level generation complicates reactive responses due to a low update rate, whereas handling it in the low-level control may cause abrupt corrections under strict timing constraints.
Another approach parametrizes the reference path with a spatial variable (e.g., arc length \cite{kabzan2020amz}) instead of time \cite{faulwasser2013optimization} often combined with \mpc formulations as \mpcc. Unlike time parameterization, \mpcc does not impose strict time constraints on path progression \cite{faulwasser2016implementation}, which suits environments with multiple constraints.

\begin{figure} [t!]
\centering
\includegraphics[width=0.48\textwidth]{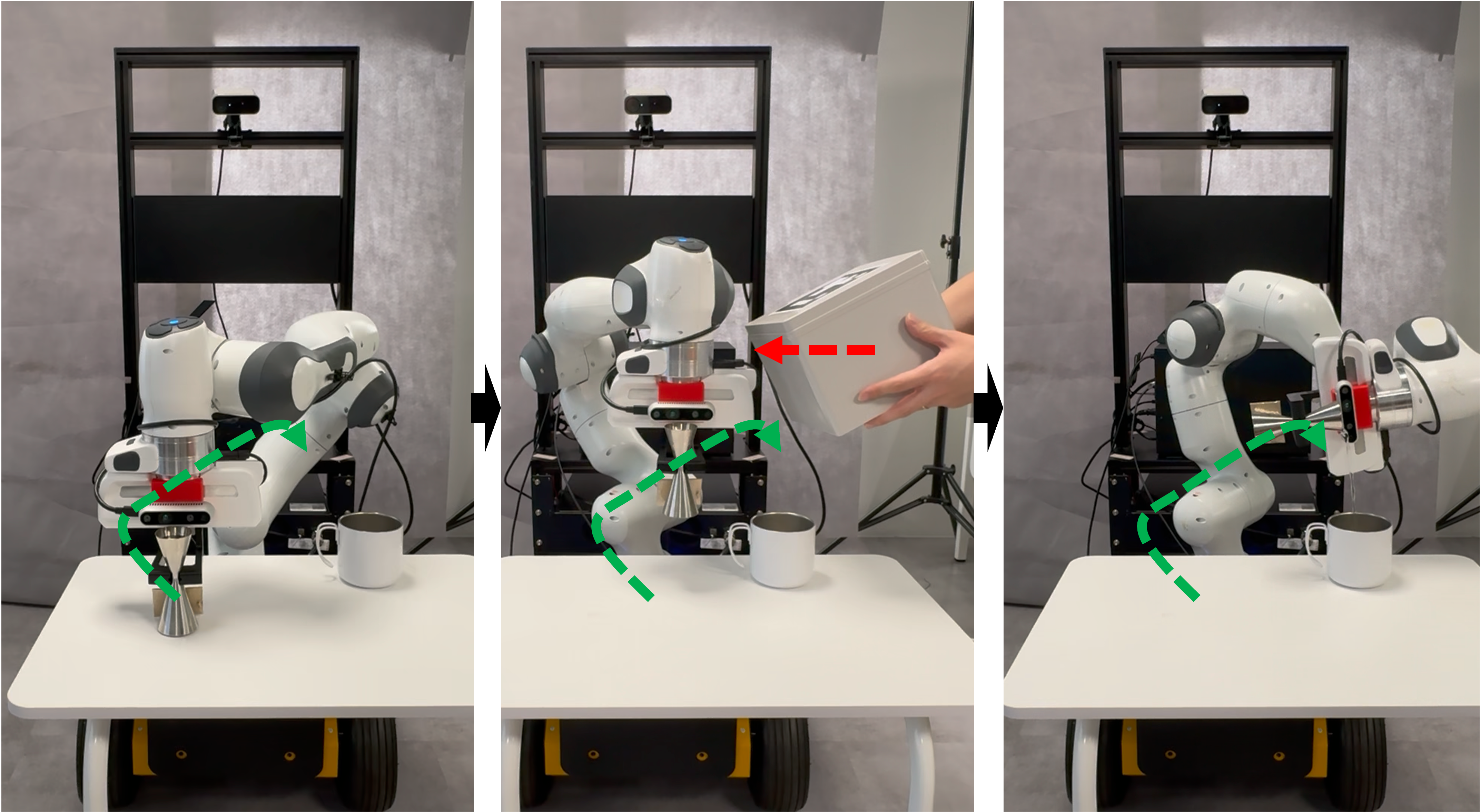}
\caption{The presented framework for RMPCC enables moving a cup containing liquid along a reference path (green line) while avoiding singularities, self-collisions, and one dynamic obstacle (red arrow) with a frequency of 100 Hz.}
\end{figure}

In this paper, \rmpcc is introduced as a robust path-following framework that leverages the advantageous features of \mpcc to operate effectively under various constraints. This method minimizes deviations from the desired reference path using a path parameter, while \cbfs handle constraints. Compared to \apf methods, \cbfs offer smoother and more stable avoidance behaviors that are minimally invasive to the nominal trajectory \cite{singletary2021comparative}. Additionally, a Jacobian-based linearization accelerates computation by about 10 times than state-of-the-art (SOTA) frameworks, making real-time operation at 100Hz feasible. Validation experiments, conducted in both simulation and real-world settings, confirm that this approach successfully avoids collisions with itself and an external obstacle, and singularities on a 7-DoF manipulator.

The remainder is organized as follows: Sec. \ref{sec:sota} reviews related path-following methods. Sec. \ref{sec:preliminaries} outlines essential preliminaries. Sec. \ref{sec:RMPCC} presents the \rmpcc framework, including collision avoidance constraints. Sec. \ref{sec:lin_jacobian} discusses the Jacobian-based linearization enabling 100 Hz operation. Sec. \ref{sec:validation} presents validation experiments and results. Sec.  \ref{sec:conclusion} concludes the paper and offers an outlook.

\section{Related Work} \label{sec:sota}
\textcolor{black}{This section reviews prior work on robot path-following, distinguishing between methods that parameterize the reference path in time and those that use a path parameter.}

\subsection{Path-following with Time Parametrization}
Sampling-based methods are popular for trajectory planning due to their probabilistic completeness \cite{elbanhawi2014sampling, karaman2011sampling}. However, their long computation times and the need for trajectory smoothing limit online applicability \cite{ferguson2006replanning}. Optimization-based approaches, such as TrajOpt \cite{schulman2014motion} and RelaxedIK \cite{rakita2018relaxedik}, enforce collision avoidance by applying no-collision constraints in discrete time and extending them to continuous time via geometric approximations. Yet, incorporating multiple constraints and nonlinearities makes these methods computationally intensive, limiting their real-time applicability.

To reduce the computational burden, \mpc was proposed for real-time trajectory-tracking frameworks \cite{nicolis2020operational, wang2024hierarchical}. Lee et al. present an efficient real-time \mpc framework that addresses hierarchical task execution and singularity avoidance, leading to smoother and more accurate task tracking at high update rates in \cite{lee2023real}. However, these approaches often struggle to maintain robust tracking performance under dynamic conditions due to the slow reactivity of high-level planning and the strict timing constraints of low-level tracking. Moreover, many \mpc frameworks developed for collision avoidance \cite{zhu2023real, liu2025flexible, nubert2020safe, wang2023fast} are mainly focused on point-to-point tasks, limiting their effectiveness in path-following scenarios.

Despite experimental validation, time-based parameterization often results in abrupt evasive maneuvers and undesired deviations from the reference path. In contrast, using a path parameter offers a more responsive and robust strategy.

\subsection{Path-following with Path Parametrization}
Unlike trajectory tracking \mpc, which enforces strict time constraints, path-parameterized path-following \mpc, commonly referred to as \mpcc, provides temporal flexibility for accurate and robust path-following in constrained environments. \textcolor{black}{For example, one study used an autonomous race car model to navigate a track under constraints \cite{kabzan2020amz}, while a further contribution applied MPCC to a quadrotor for following complex paths \cite{romero2022model}. Moreover, MPCC has been shown to enable manipulators to accurately track reference paths and handle disturbances \cite{van2016path, faulwasser2016implementation}. However, these approaches consider only position tracking, and \cite{astudillo2022position} assumes negligible orientation errors. 
In addition, these methods focus on joint limits or tracking-error bounds without explicitly incorporating safety constraints like collision avoidance}. Consequently, their applicability in complex environments is limited.
The BoundMPC framework in \cite{oelerich2024boundmpc} utilizes Lie theory to address both position and orientation errors while adhering to Cartesian constraints. \textcolor{black}{However, due to the significant nonlinearity in its formulation, the method runs at a relatively low control frequency of around 10 Hz. This limited update rate makes it difficult to handle dynamic obstacles or rapidly changing environments in real time.}
\textcolor{black}{Moreover, an \apf is integrated into its cost function; according to \cite{singletary2021comparative}, APFs can lead to intrusive control actions during obstacle avoidance when compared to methods that employ \cbfs.}

In contrast, our framework can handle dynamic changes, obstacles, and self-collisions \textcolor{black}{by \cbfs} with high control frequency of 100 Hz. To this end, the contributions of this paper can be summarized as follows: 
\begin{enumerate}
    \item The \rmpcc framework introduces a novel formalism to incorporate the \cbfs, enabling the avoidance of singularities, self-collisions as well as external collisions during path-following operation. 
    \item The linearization via a Jacobian combined with the Gauss-Newton Hessian approximation yields a frequency of 100 Hz, thereby outperforming SOTA methods by a factor of 10. 
\end{enumerate}

\section{Preliminaries} \label{sec:preliminaries}
\subsection{Model Predictive Contouring Control (MPCC)} \label{MPCC}
\begin{figure}[t!]
    \centering
    \includegraphics[width=0.4\textwidth]{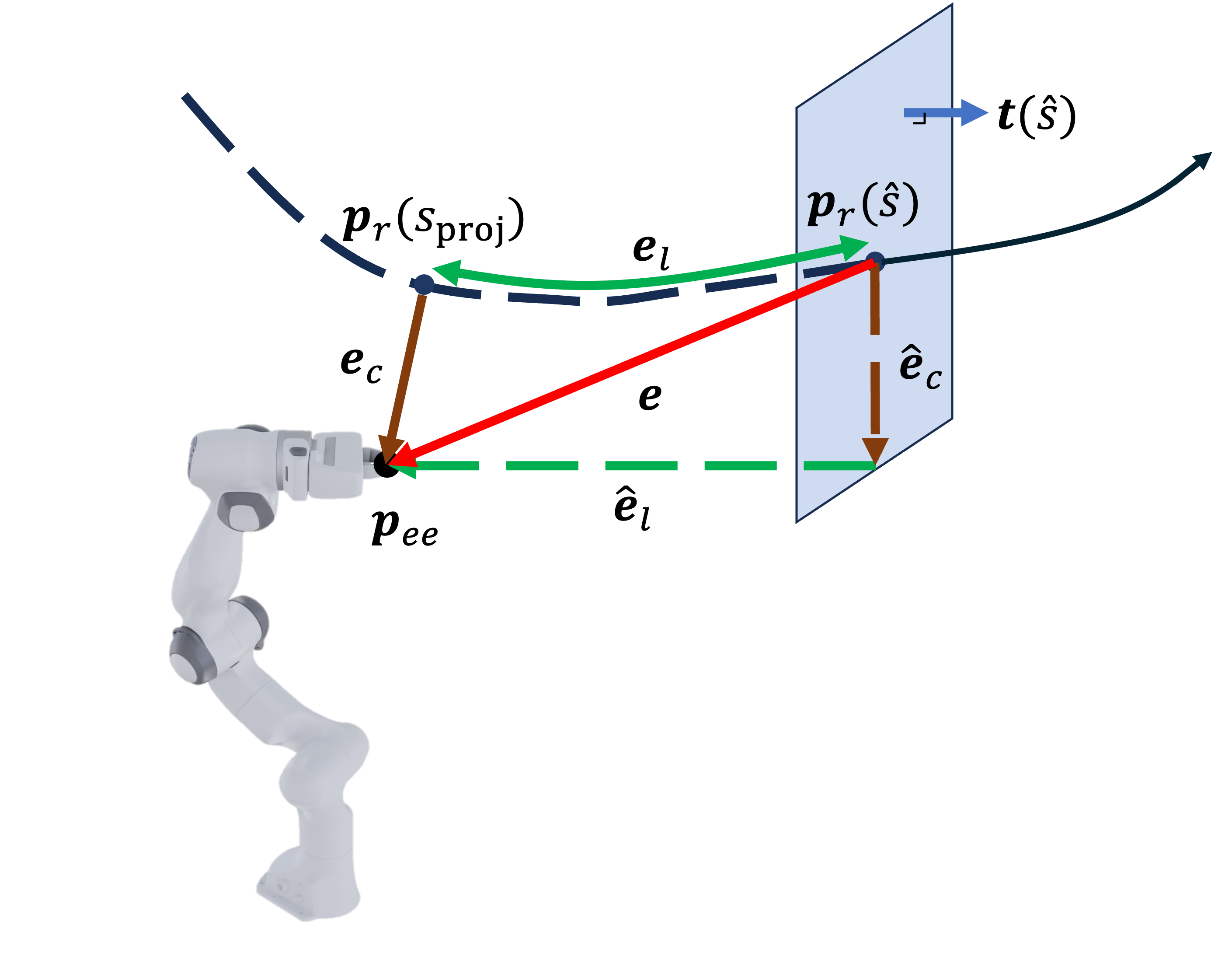}
    \caption{Visualization of the lag error $\be_l$ (green line), contouring error $\be_c$ (brown line), and overall error $\be$  (red line) between the robot $\bp_{ee}$ and the reference path $\bp_{r}$. The projected path parameter $s_{\text{proj}}$ is defined to minimize the contouring error. Approximated errors $\hat{\be}_c$ and $\hat{\be}_l$ are depicted as dotted lines.}
    \label{fig:mpcc_mani}
\end{figure}

In \mpcc, the path is parameterized by a path parameter $s \in \R$ such that $\bp_{r}(s) = [x_{r}(s) \ y_{r}(s) \ z_{r}(s)]^{\top}$. To accurately follow the path, contouring error $\be_c$, which represents the Euclidean error between the position of the robot $\bp_{ee}$ and the projected position on the reference path $\bp_{r}(s_{\text{proj}})$  (see Fig. \ref{fig:mpcc_mani}), should be minimized. 
Since obtaining $s_{\text{proj}}$ directly is difficult, an approximation $\hat{s}$ is introduced with dynamics 
\begin{equation}
\begin{aligned}
    \hat{s}_{k+1} &= \hat{s}_k + v_{s,k}\Delta t + \frac{1}{2}\dot{v}_{s,k}\Delta t^2 \\
    v_{s,k+1} &= v_{s,k} + \dot{v}_{s,k}\Delta t
\end{aligned}
\end{equation}
where $\dot{v}_{s,k}$ acts as a virtual control input in the \ocp. By appropriately adjusting $\dot{v}_{s,k}$, $\hat{s}$ can converge to $s_{\text{proj}}$. The resulting discrepancy between $s_{\text{proj}}$ and $\hat{s}$ is defined as the lag error $\be_l$. Since $s_{\text{proj}}$ is not known within the \ocp, both $\be_c$ and $\be_l$ are approximated as \cite{romero2022model}
\begin{equation} \label{contouring error}
    \be=\bp_{r}(\hat{s}) - \bp_{ee}, \;
    \hat{\be}_l = \left( \bt(\hat{s})^{\top} \be \right)  \bt(\hat{s}), \;
    \hat{\be}_c = \be - \hat{\be}_l. 
\end{equation}
Here, $\bt(\hat{s}) \in \R^3$ represents the tangent vector of the path at $\bp_{r}(\hat{s})$. As the approximated lag error $\hat{\be}_l$ approaches zero, the approximated contouring error $\hat{\be}_c$ converges to the actual contouring error $\be_c$ \cite{romero2022model}. Therefore, \mpcc achieves accurate path-following by minimizing $\hat{\be}_c$ and $\hat{\be}_l$. For the remainder of this paper, the hat operator will be omitted for simplicity.

\subsection{Control Barrier Function (CBF)} \label{CBF}
Consider a dynamical system $\dot{\bx} = f(\bx, \bu)$, where $\bx \in \mathbb{R}^{n_x}$ is the state and $\bu \in \R^{n_u}$ is the input. A continuously differentiable function $h: \R^{n_x} \to \R$ quantifies safety (e.g., self-minimum distance), with the safe set defined as
$C = \{\, \bx \in \mathbb{R}^{n_x} \mid h(\bx) \geq 0 \}$. If $\bx(t) \in \mathcal{C}$ holds for all $t$, then the set $C$ is classified as \emph{forward invariant} and the system is considered safe \cite{ames2019control}.
A CBF enforces safety by modifying a nominal control input $k(\bx, t)$ while regulating the time derivative of $h$ in \qp
\begin{equation} \label{cbf_qp}
\begin{aligned}
    \min_{\bu} \frac{1}{2}\|\bu-\bm{k}(\bx, t)\|_2^2 \; ; \;\;\;  \dot{h}(\bx, \bu) \geq -\gamma( h(\bx)),
\end{aligned}
\end{equation}
where $\gamma: \R \rightarrow \R$ is a class $\mathcal{K}_{\infty}$ function. For a system with generalized coordinates $\bz \in \R^{n_z}$, velocities $\dot{\bz} \in \R^{n_z}$, and $h$ depending only on $\bz$, the time derivative of $h$ is
    $\dot{h}(\bz, \dot{\bz}) = \nabla_{\bz} h(\bz)^\top \dot{\bz}$.
In this work, \rbf \cite{grandia2019feedback} is adopted for $\gamma$, defined as:
\begin{equation}
    \renewcommand{\arraystretch}{1.0}
    \RBF(h)=\left\{ \begin{array}{rcl}
    -\log(h+1) & \text{for} & h \geq \delta, \\ 
    \beta(h;\delta) & \text{for} & h < \delta,
    \end{array}\right.
\end{equation}
where $\beta(\cdot)$ is a quadratic function and $\gamma(h) = -\RBF(h)$. Thus, the constraint of \eqref{cbf_qp} is reformulated as
\begin{equation} \label{cbf_form}
    c_h(\bz) = \RBF(h(\bz)) - \nabla_{\bz} h(\bz)^\top \dot{\bz} \leq 0.
\end{equation}

\subsection{SO(3) Manifold} \label{SO3 manifold}
The path a robot follows \textcolor{black}{includes} both translation and orientation, which is represented by a rotation matrix $\bR \in \operatorname{SO}(3)$. The tangent space at the identity of $\operatorname{SO}(3)$ is $\mathfrak{so}(3)$, consisting of skew-symmetric matrices generated by vectors in $\R^3$. The mappings $(\cdot)^{\wedge}:\R^3 \to \mathfrak{so}(3)$ and $(\cdot)^{\vee}:\mathfrak{so}(3)\to\R^3$ convert between vector and skew-symmetric forms. \textcolor{black}{For mapping between $\operatorname{SO}(3)$ and $\R^3$ directly,} the exponential and logarithmic maps are \textcolor{black}{defined} according to \cite{forster2016manifold}:
\begin{equation} \label{Log and Exp}
\begin{aligned}
    & \Exp: \R^3 \rightarrow \operatorname{SO}(3) \ ; \ \bm{\phi} \mapsto \exp(\bm{\phi}^{\wedge}), \\
    & \Log: \operatorname{SO}(3)\rightarrow \R^3 \ ; \ \bR \mapsto \log(\bR)^{\vee},
\end{aligned}
\end{equation}
where \textcolor{black}{$\bm{\phi} \in \R^3$ is a rotation vector \cite{forster2016manifold}.} After applying a small rotation $\delta \bm{\phi} \in \R^3$ to $\bR_{\text{init}}$,
\begin{equation} \label{Exp R}
    \bR = \Exp(\delta \bm{\phi}) \bR_{\text{init}}.
\end{equation}
First-order approximations of the logarithmic map \cite{forster2016manifold} give
\begin{align} \label{Exp Log Jacobian}
    \Log\!\bigl(\Exp(\bm{\phi}) \Exp(\delta \bm{\phi})\bigr) \approx \bm{\phi} + \bJ_r^{-1}(\bm{\phi})\delta \bm{\phi}.
\end{align}
Here, $\bJ_r(\bm{\phi})^{-1}$ is the inverse of the right Jacobian of $\operatorname{SO}(3)$:
\begin{equation} \label{Jr}
    \bJ_r(\bm{\phi})^{-1} = \mathbf{I} + \frac{1}{2} \bm{\phi}^{\wedge} + \left( \frac{1}{\|\bm{\phi}\|^2} + \frac{1+\cos{\|\bm{\phi}\|}}{2\|\bm{\phi}\| \sin{\|\bm{\phi}\|}} \right) \left( \bm{\phi} ^{\wedge} \right)^2 .
\end{equation}
Lastly, the exponential map satisfies the property \cite{forster2016manifold}:
\begin{equation} \label{exp property}
    \Exp(\bm{\phi}) \bR = \bR \Exp\left(\bR^{\top} \bm{\phi}\right).
\end{equation}
\begin{figure*}[!t]
    \centering
    \includegraphics[width=1\textwidth]{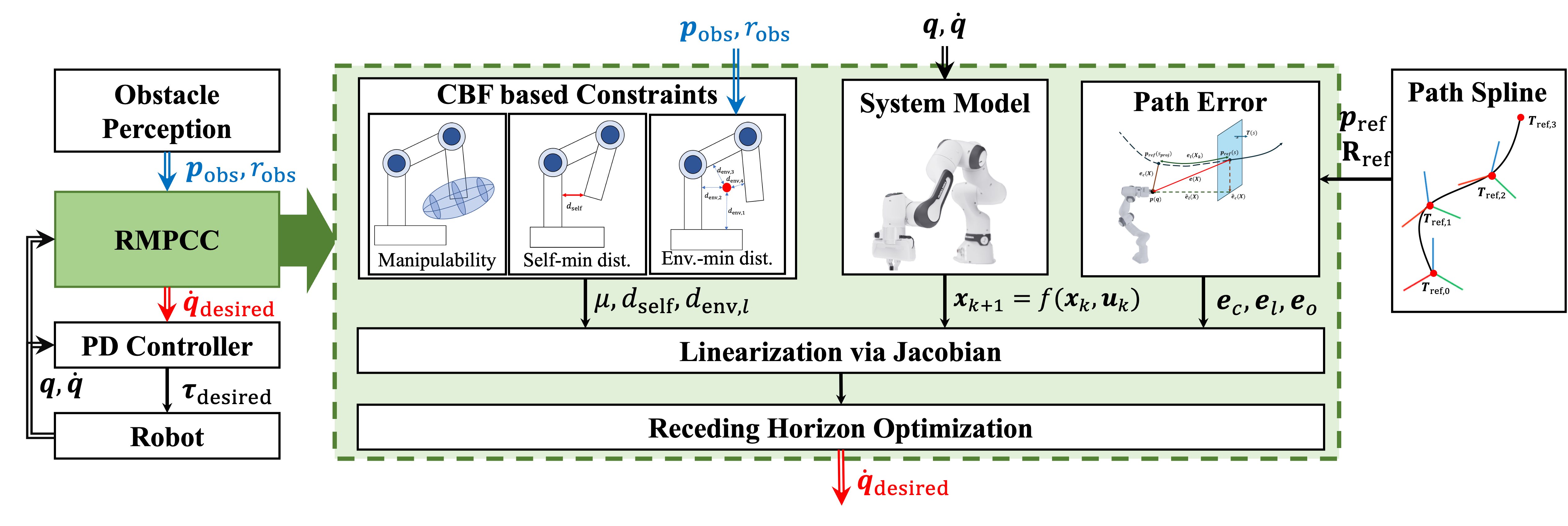}
    \caption{Schematic overview of our \rmpcc framework. The system model, path spline, and \cbfs are integrated into a receding-horizon optimization. A Jacobian-based linearization and Gauss-Newton Hessian approximation enable rapid solution at 100 Hz ensuring real-time collision and singularity avoidance.}
    \label{fig:schematic diagram}
\end{figure*}

\section{Reactive MPCC (RMPCC) Formulation} \label{sec:RMPCC}
In this section, we introduce the \rmpcc formulation, a $\operatorname{SE}(3)$ path-following strategy  that accounts for kinematic constraints. Fig.~\ref{fig:schematic diagram} illustrates the overall framework. Sec.~\ref{sec:path spline} presents a path parameterization using via-points. \textcolor{black}{The system model is presented in Sec.~\ref{sec:system model}, and} cost terms for minimizing path-following error and constraints for robot system stability are addressed in Sec. \ref{sec: cost} and Sec.\ref{sec: constraints}.

\subsection{Path Spline} \label{sec:path spline}
\textcolor{black}{Unlike the arc length parameterization in \cite{kabzan2020amz, romero2022model}, \rmpcc employs a normalized parameter $s \in [0, 1]$ since representing orientation using arc length is challenging.} Given $P$ via-points $T_{\text{ref}, 1}, \dots, T_{\text{ref}, P}$ \textcolor{black}{with equally spaced $s_i$ values from 0 to 1, the position is splined using continuously differentiable third-order splines $\bm{\rho}_i(s)$, and the orientation is interpolated using $\bm{\phi}_i(s)$ combined with the exponential mapping operation $\operatorname{Exp}(\cdot)$ in \eqref{Log and Exp}.} Each reference pose $T_{\text{ref}, i} \in \operatorname{SE}(3)$ \textcolor{black}{contains} both the position $\bp_{\text{ref},i} \in \mathbb{R}^3$ and the orientation $\bR_{\text{ref}, i} \in \operatorname{SO}(3)$.

\begin{equation} \label{position spline}
\renewcommand{\arraystretch}{0.5}
\bp_{r}\left(s\right)=\left\{\begin{array}{cl}
\bm{\rho}_1\left(s\right) 
& s_1 \leq s \leq s_2 \\
\vdots & \\
\bm{\rho}_{P-1}\left(s\right) 
& s_{P-1} \leq s \leq s_{P}
\end{array}\right.
\end{equation}

\begin{equation} \label{orientation spline}
\renewcommand{\arraystretch}{0.5}
\begin{aligned}
    & \bR_{r}\left(s\right)=\left\{
        \begin{array}{cl}
        \bR_{\text{ref}, 1} \Exp(\bm{\phi}_{1}(s)) 
        & s_1 \leq s \leq s_2 \\
        \vdots & \\
        \bR_{\text{ref}, P-1} \Exp(\bm{\phi}_{P-1}(s)) 
        & s_{P-1} \leq s \leq s_{P}
        \end{array} \right.
    \\
\end{aligned}
\end{equation}
Here, $\bm{\phi}_i(s)= \alpha_i(s) \Log(\bR_{\text{ref}, i}^{\top} \bR_{\text{ref},i+1})$ and $\alpha_i: \R \rightarrow \R$ is cubic polynomial function which satisfies $\alpha_i(s_i)=\alpha^{\prime}_i(s_i)=\alpha^{\prime}_i(s_{i+1})=0$ and $\alpha_i(s_{i+1})=1$.

\subsection{System Model} \label{sec:system model}
\rmpcc controls an $n$-DoF manipulator to follow a reference path. The state is defined by the joint angles $\bq = [q_1 \ \dots \ q_n]^{\top}\in\R^n$, and the control inputs are the joint velocities $\dot{\bq} = [\dot{q}_1 \ \dots \ \dot{q}_n]^{\top}\in\R^n$. The end-effector’s position $\bp_{ee}\in\R^3$ and orientation $\bR_{ee}\in\operatorname{SO}(3)$ are obtained via forward kinematics, while its linear and angular velocities $\bm{v}_{ee},\bm{\omega}_{ee}\in\R^3$ are computed using the Jacobian
\begin{equation} \label{jacobian}
\renewcommand{\arraystretch}{1.0}
\begin{aligned}
    \begin{bmatrix}
    \bm{v}_{ee}  \\ \bm{\omega}_{ee}
    \end{bmatrix}
     =
    \begin{bmatrix}
    \bJ_{\text{posi}} (\bq)  \\ \bJ_{\text{ori}} (\bq)
    \end{bmatrix} 
    \dot{\bq}
    =
    \bJ_{ee}(\bq) \dot{\bq}. 
\end{aligned}
\end{equation}

In \rmpcc, a model of the path parameter $s$ explained in Sec. \ref{MPCC} must be taken into account. The system state $\bx$ and control input $\bu$ are defined as
\begin{equation} \label{model}
    \bx=[\bq \ s \ v_s]^\top \in \R^{n+2},
    \quad \bu = [\dot{\bq} \ \dot{v}_s]^\top \in \R^{n+1}
\end{equation}
The continuous-time system model can be expressed by
\begin{equation}
\renewcommand{\arraystretch}{1.0}
\begin{aligned}
    \dot{\bx} = 
    f_c(\bx, \bu) & =
    \begin{bmatrix}
        \bm{0} & \bm{0} & \bm{0} \\
        \bm{0} & 0  & 1 \\ \bm{0} & 0 &0
    \end{bmatrix}
    \begin{bmatrix}
        \bq \\
        s \\
        v_{s}
    \end{bmatrix}
    +
    \begin{bmatrix}
        \mathbf{I} & \bm{0} \\ \bm{0} & 0 \\
        \bm{0} & 1
    \end{bmatrix}
    \begin{bmatrix}
         \dot{\bq} \\
        \dot{v}_{s}
    \end{bmatrix}
     \\ &= A_c \bx + B_c \bu.
\end{aligned}
\end{equation}
Since $f_c$ is a linear time-invariant system, the discretized model becomes
\begin{equation} \label{system model}
    \bx_{k+1} = e^{A_c \Delta t} \bx_k + \textcolor{black}{(e^{A_c \Delta t})^{-1}(A_c- \mathbf{I}) B_c \bu_k}
     = f(\bx_k, \bu_k)
\end{equation}
where $\Delta t$ is the sample time of the robot system.

\subsection{Cost Function} \label{sec: cost}
The cost of the receding-horizon \ocp is defined as the sum of stage costs over the prediction horizon
\begin{equation}
    J(\bx_{1:N}, \bu_{1:N-1}) = L_f(\bx_N) + \sum_{k=1}^{N-1}{L(\bx_k, \bu_k)}
\end{equation}
The stage cost $L(\bx, \bu)$ and the terminal cost $L_f(\bx)$ are formulated as the sum of three sub-cost functions: the cost for contouring and lag error $L_{\text{cont}}$, orientation error $L_{\text{ori}}$ and control input regularization $L_{\text{input}}$.

\subsubsection{Contouring and Lag Error} \label{sec331}
To ensure that the robot follows the reference path, the contouring error $\be_c$ and the lag error $\be_l$, calculated using   \eqref{contouring error}, must be minimized by defining
\begin{equation} \label{L_contour}
\begin{aligned}
    &L_{\text{cont}}(\bx) = \textcolor{black}{w}_{c}\| \be_{c}(\bx) \|_2^2 + \textcolor{black}{w}_{l}\| \be_{l}(\bx) \|_2^2 + \textcolor{black}{w}_{v_s}(v_{\text{desired}} - v_{s})^2. \\
\end{aligned}
\end{equation}
Here, $\textcolor{black}{w}_{c}, \textcolor{black}{w}_{l}, \textcolor{black}{w}_{v_s} \in \R^+$ are weighing scalars for contouring, lag and velocity of path parameter error, respectively. And $v_{\text{desired}} \in \R^+$ is desired velocity of path parameter.

\subsubsection{Orientation Error} \label{sec332}
For $L_{\text{ori}}$, $\be_o(\bx)$ is defined by the difference between two rotation matrices $\bR_{ee}$ and $\bR_{r}$.
\begin{equation} \label{L_ori}
\begin{aligned}
    L_{\text{ori}}(\bx) &= \textcolor{black}{w}_{o}\|\be_o(\bx)\|_2^2 \\
    \be_o(\bx) &= \Log(\bR_{r}(s)^{\top} \bR_{ee}(\bq) )\in \R^3
\end{aligned}
\end{equation}
where $\textcolor{black}{w}_{o} \in \R^+$ is a weighing scalar for the orientation error and the operator $\Log(\cdot)$ is defined in   \eqref{Log and Exp}.

\subsubsection{Control Input Regularization} \label{sec334}
To prevent excessive manipulator movements, $L_{\text{input}}$ is defined as: 
\begin{equation} \label{cost_input}
    L_{\text{input}}(\bu_k, \bu_{k-1})= \textcolor{black}{w}_{\dot{q}}\|\dot{\bq}_k\|_2^2+ \textcolor{black}{w}_{\Delta \dot{q}}\| \Delta \dot{\bq_k} \|_2^2+ \textcolor{black}{w}_{\dot{v}_s}\dot{v}_{s,k}^2
\end{equation} 
where $\textcolor{black}{w}_{\dot{q}}, \textcolor{black}{w}_{\Delta \dot{q}}, \textcolor{black}{w}_{\dot{v}_s} \in \R^+$ represent weighting scalars for joint velocity, joint velocity differences, and path parameter acceleration regularization, respectively. Additionally, $\Delta \dot{\bq}_k$ is defined as $\bm{0} \in \mathbb{R}^n$ for $k = 1$ and as $\dot{\bq}_k - \dot{\bq}_{k-1}$ otherwise.

\subsection{Constraints} \label{sec: constraints}
To ensure safe operation, three constraints, which are singularity, self-collision, and environment-collision avoidance, are imposed, following the formulation in \eqref{cbf_form}. Next, their corresponding barrier functions are defined.

\subsubsection{Singularity Avoidance} \label{sec342}
To avoid singularities, a constraint is imposed to ensure that the manipulability index $\mu$ remains above a certain threshold.
\begin{equation} \label{constraint_sing}
    h_{\text{sing}}(\bq) = \textcolor{black}{\mu}(\bq) - \epsilon_{\text{sing}} \geq 0
\end{equation}
The manipulability index $\textcolor{black}{\mu}(\bq): \mathbb{R}^n \rightarrow \mathbb{R}$ quantifies how well-conditioned a manipulator is for achieving arbitrary end-effector velocities \cite{haviland2020purely}.
\begin{equation}
    \textcolor{black}{\mu}(\bq)=\sqrt{\det{[\bJ_{ee}(\bq)\bJ_{ee}(\bq)^\top]}}
\end{equation}

\subsubsection{Self-collision Avoidance} \label{sec343}
To prevent self-collisions, the minimum distance between the links of the manipulator $d_{\text{self}} \in \R$ must stay above a threshold: 
\begin{equation} \label{constraint_selfcol}
    h_{\text{self}}(\bq) = d_{\text{self}}(\bq) - \epsilon_{\text{self}} \geq 0.
\end{equation}
The minimum distance $d_{\text{self}}$ is predicted using a \nn model. Details of the \nn model are provided in Appendix \ref{app.A}.

\subsubsection{Environment-collision Avoidance} \label{sec344}
External collision avoidance requires the minimum distances between robot links and an obstacle $ d_{\text{env}, l} $ to stay above a threshold. For a robot with $L$ links, this condition is expressed as:
\begin{equation} \label{constraint_envcol}
\begin{aligned}
     h_{\text{env}, l}&(\bq, \bp_{\text{obs}}) \\ &= d_{\text{env}, l}(\bq, \bp_{\text{obs}}) - r_{\text{obs}} - \epsilon_{\text{env}} \geq 0, \; 1\leq l \leq L
\end{aligned}
\end{equation}
where $ d_{\text{env}, l} $ denotes the distance between the $ l $-th link of the robot and the center of a spherical obstacle with its radius $r_{\text{obs}}$. To compute $ {d}_{\text{env}, l} $, a neural joint space implicit signed distance function (Neural-JSDF), proposed in \cite{koptev2022neural}, is used. This method uses a \nn which takes the position of the external obstacle $\bp_{\text{obs}} \in \R^3$ and the joint angles of the robot $\bq$ as inputs and predicts the minimum distance $ {d}_{\text{env}, l} $.

Finally, the \ocp of \rmpcc can be formulated as
\begin{equation} \label{mpcc ocp}
\begin{aligned}
\min_{\substack{\bx_{1:N}\\ \bu_{1:N-1}}}& 
 L_f(\bx_N) + 
\sum_{k=1}^{N-1} 
    L(\bx_k, \bu_k, \bu_{k-1}) \\[1ex]
\text{subject to} \quad
& \bx_1 = \bx_{\text{initial}}, \\
& \bx_{k+1} = f(\bx_k, \bu_k), \\
& \bx_{\text{min}} \leq \bx_k \leq \bx_{\text{max}}, 
\quad \bu_{\text{min}} \leq \bu_k \leq \bu_{\text{max}}, \\
& \RBF\bigl(h_{\text{sing}}(\bq_k)\bigr) 
    \leq \nabla_{\bq}h_{\text{sing}}(\bq_k)^{\top}\dot{\bq}_k, \\
& \RBF\bigl(h_{\text{self}}(\bq_k)\bigr) 
    \leq \nabla_{\bq}h_{\text{self}}(\bq_k)^{\top}\dot{\bq}_k, \\
& \RBF\bigl(h_{\text{env}, l}(\bq_k, \bp_{\text{obs}})\bigr) 
    \leq \nabla_{\bq}h_{\text{env}, l}(\bq_k)^{\top}\dot{\bq}_k, \\
& \quad \quad \quad \quad \quad   1 \leq k \leq N-1, \;  1 \leq l \leq L
\end{aligned}
\end{equation}
where $N$ denotes the prediction horizon length.

\section{Linearization via Jacobian} \label{sec:lin_jacobian}
To solve the nonlinear \ocp  \eqref{mpcc ocp} in real-time, a \sqp algorithm in \cite{jordana2023stagewise} \textcolor{black}{is adopted}. \textcolor{black}{To do} that, Hessian and gradient of the stage cost $L(\bx, \bu)$ and the constraints $c_h(\bx, \bu)$ in \eqref{mpcc ocp} must be calculated.

\subsection{Cost Function} \label{sec41}
Since calculating the second-order derivatives of $\be_c, \be_l, \be_o$ is computationally expensive \cite{chen2011hessian}, the Gauss-Newton Hessian approximation is employed, which neglects second-order derivative terms. Hence, the computation is reduced to the determination of the Jacobians of $\be_c, \be_l, \be_o$.

\subsubsection{Contouring and Lag error} \label{sec411}
Before deriving the Jacobians for $\be_c$ and $\be_l$, the Jacobians for the total error $\be(\bx)$ and the tangent vector $\bt(s)$ of the reference path need to be computed. The Jacobian of the total error is calculated as:
\begin{equation} \label{Jacobian error}
\begin{aligned}
    \frac{\partial \be}{\partial \bx}  
    &= \frac{\partial \bp_{ee}}{\partial \bx} - \frac{\partial \bp_{r}}{\partial \bx} \\
    &= \begin{bmatrix} 
    \frac{\partial \bp_{ee}}{\partial \bq} & \frac{\mathrm{d} \bp_{ee}}{\mathrm{d} s}  & \frac{\mathrm{d} \bp_{ee}}{\mathrm{d} v_s} \end{bmatrix} 
    -
    \begin{bmatrix} 
    \frac{\partial \bp_{r}}{\partial \bq} & \frac{\mathrm{d} \bp_{r}}{\mathrm{d} s}  & \frac{\mathrm{d} \bp_{r}}{\mathrm{d} v_s} 
    \end{bmatrix}  \\
    &=\begin{bmatrix} 
    \bJ_{\text{posi}}  & -\bm{t} & \bm{0}
    \end{bmatrix} 
\end{aligned}
\end{equation}
Next, the Jacobian of $\bm{t}$ with respect to $\bx$ is derived as:
\begin{equation} \label{Jacobian tangent}
    \frac{\partial \bm{t}}{\partial \bx} = 
    \begin{bmatrix} 
    \frac{\partial \bm{t}}{\partial \bq}
     & 
    \frac{\mathrm{d} \bm{t}}{\mathrm{d} s}  & \frac{\mathrm{d} \bm{t}}{\mathrm{d} v_s}  \end{bmatrix}  = \begin{bmatrix} \bm{0} & \bm{n} & \bm{0} \end{bmatrix}
\end{equation}
where $\bm{n}$ is the derivative of the tangent vector, i.e., the path's normal direction. Using   \eqref{Jacobian error} and   \eqref{Jacobian tangent}, the Jacobian of the lag error $\be_l$ is derived as:
\begin{equation} \label{Jacobian lag error}
    \frac{\partial \be_l}{\partial \bx} 
    = (\bm{t}^\top \be)\frac{\partial \bm{t}}{\partial \bx} 
    + \bm{t} \be^\top \frac{\partial \bm{t}}{\partial \bx}  
    + \bm{t} \bm{t}^\top \frac{\partial \be}{\partial \bx}  \\
\end{equation}
Finally, since $\be = \be_c + \be_l$ by definition, the Jacobian of the contouring error is:
\begin{equation}
    \frac{\partial \be_c}{\partial \bx}  =\frac{\partial \be}{\partial \bx}  - \frac{\partial \be_l}{\partial \bx}.
\end{equation}

\subsubsection{Orientation error} \label{sec412}

The Jacobian of $\be_o$ is given by
\begin{equation} \label{jacoian_eo}
\begin{aligned}
    \frac{\partial \be_o}{\partial \bx} 
    &=
    \begin{bmatrix} 
    \frac{\partial \be_o}{\partial \bq} & \frac{\partial \be_o}{\partial s} & \frac{\partial \be_o}{\partial v_s}
    \end{bmatrix} \\
    &\approx
    \begin{bmatrix} 
    \bJ_r^{-1}\left ({\be}_o\right) {\bR}_{ee}^\top {\bJ}_{\text{ori}} & -\bJ_r^{-1}\left ({\be}_o\right ){\bR}_{ee}^\top \; {\bm{\phi}^{\prime}_{i}} & \bm{0}
    \end{bmatrix}
\end{aligned}
\end{equation}
where $\bJ^{-1}_r(\cdot)$ defined in \eqref{Jr}, and ${\bm{\phi}_{i}^{\prime}}$ is defined as $\tfrac{d\bm{\phi}_i}{ds}$ with $s_i \leq s \leq s_{i+1}$ from \eqref{orientation spline}. The derivation of \eqref{jacoian_eo} is provided in Appendix \ref{app.B}.

\subsection{Constraints} \label{sec42}
The gradient of $c_h(\bx, \bu)$ with respect to $\bx$ is expressed as
\begin{equation}
\begin{aligned}
    \nabla_{\bx} c_h(\bx, \bu) 
    &  \stackrel{(a)}{\approx} \frac{\mathrm{d} \RBF(h)}{\mathrm{d} h} \nabla_{\bx} h(\bq) \\
    & =
    \begin{bmatrix}
         \frac{\mathrm{d} \RBF(h)}{\mathrm{d} h} \nabla_{\bq} h(\bq)^\top & 0 & 0
    \end{bmatrix}^\top,
\end{aligned}
\end{equation}
where (a) the term $\nabla_{\bx} \left( \textcolor{black}{\nabla_{\bx}} h(\bq)^\top \dot{\bq} \right)$ is omitted to reduce the computational complexity. Similarly, the gradient of $c_h(\bx, \bu)$ with respect to $\bu$ is given by:
\begin{equation}
    \nabla_{\bu} c(\bx, \bu) = 
    \begin{bmatrix}
        -\nabla_{\bq} h(\bq)^\top & 0
    \end{bmatrix}^\top
\end{equation}

\section{Validation Experiments} \label{sec:validation}
\subsection{Experimental Setup}

\begin{figure}[!t]
    \centering
    \includegraphics[width=0.45\textwidth]{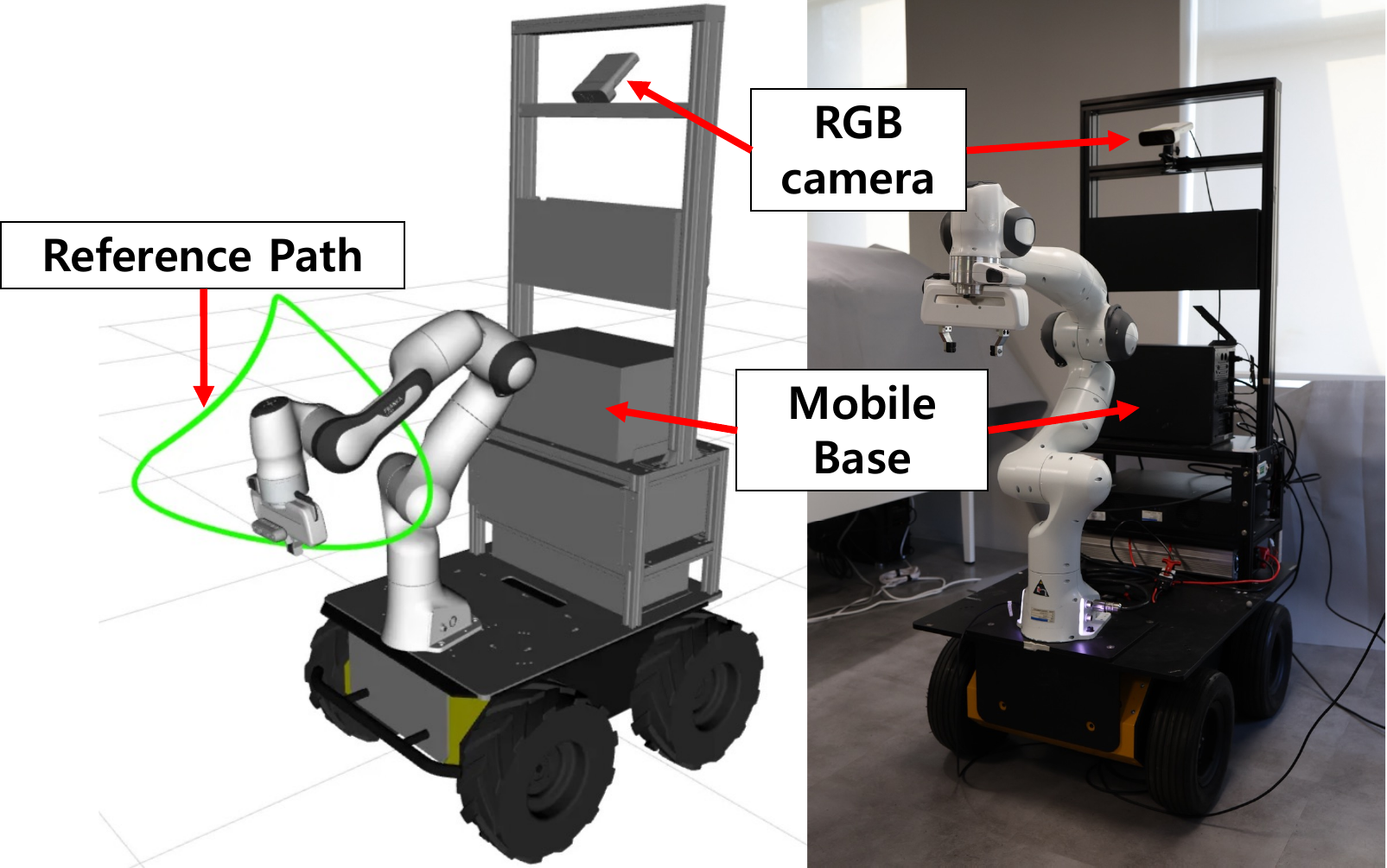}
    \caption{Experimental setup of a 7-DoF Panda manipulator on a mobile base with an RGB camera for obstacle perception in simulation (left) and real-world (right). The robot follows a 3D lemniscate path.}
    \label{fig:hardware setup}
\end{figure}

\begin{table}[h!!]
    \renewcommand{\arraystretch}{1.0}
    \centering
    \caption{\rmpcc Parameters.}
    \label{tab:parameters}
    \begin{tabular}{c | c || c | c || c | c}
    \hline
    \textbf{Param} & \textbf{Value} & \textbf{Param} & \textbf{Value} & \textbf{Param} & \textbf{Value}\\
    \hline
    $N$        & 10      & $q_o$                 & 100   & $\epsilon_{\text{sing}}$ & 0.018 \\
    $\textcolor{black}{w}_c$      & 500     & $\textcolor{black}{w}_{\dot{q}}$         & 0.002 & $\epsilon_{\text{self}} \; \text{[cm]}$ & 1 \\
    $\textcolor{black}{w}_l$      & 100     & $\textcolor{black}{w}_{\Delta \dot{q}}$  & 10    & $\epsilon_{\text{env}} \; \text{[cm]}$  & 1 \\
    $\textcolor{black}{w}_{v_s}$  & 2       & $\textcolor{black}{w}_{\dot{v}_s}$       & 0.1   & $v_{\text{desired}}$     & 0.05  \\
    \hline
    \end{tabular}
\end{table}

To validate the \rmpcc framework, the 7-DoF Panda manipulator (\textit{Franka Emika}) was mounted on a mobile base (not used during experiments) as shown in Fig. \ref{fig:hardware setup}. OSQP\footnote{https://osqp.org} was used as the \qp solver, and all experiments were conducted at 100 Hz. Parameter values are listed in Table \ref{tab:parameters}.

\begin{figure*}[t!!]
    \centering
    \begin{subfigure}{0.49\textwidth}
        \centering
        \includegraphics[width=1\textwidth]{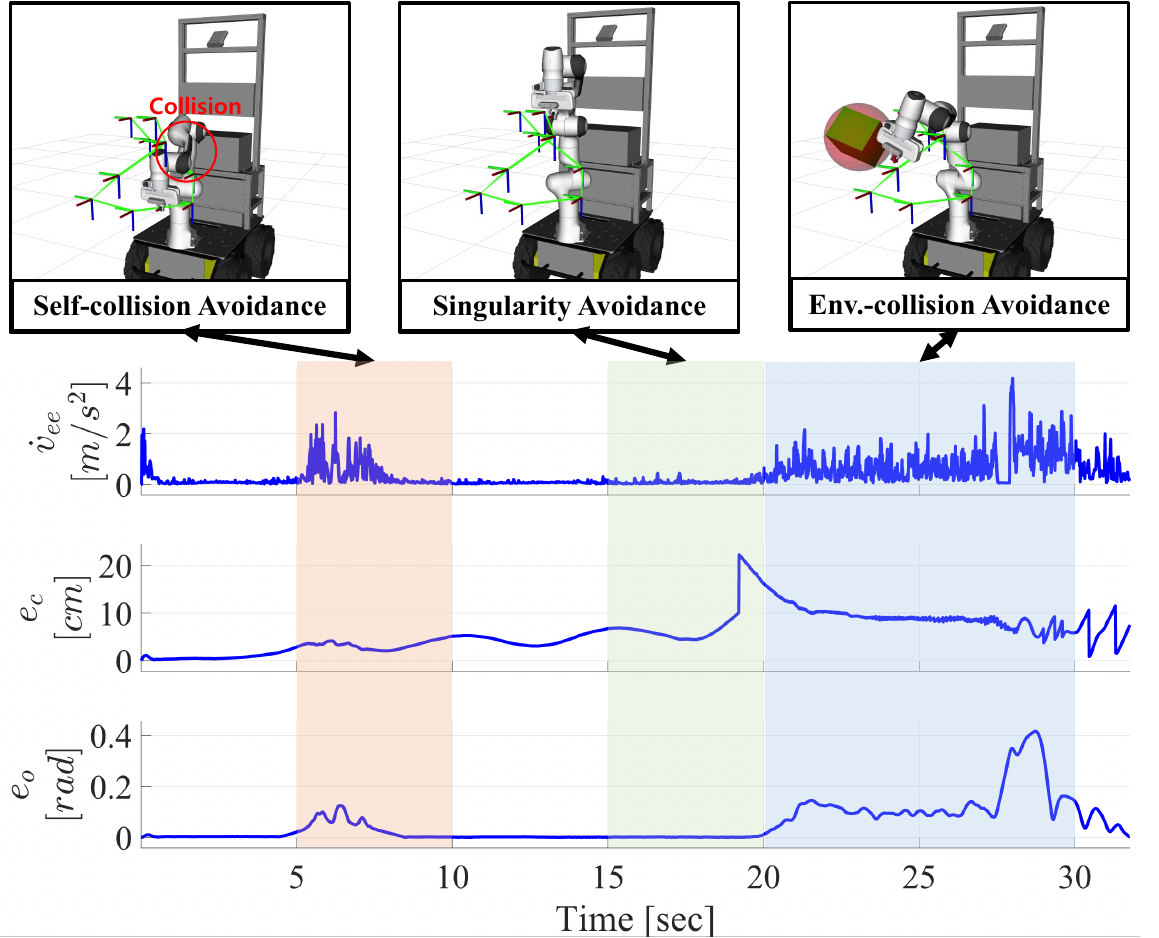}
        \caption{Result of TT-MPC}
        \label{fig:performance_ttmpc}
    \end{subfigure}
    \hfill
    \begin{subfigure}{0.49\textwidth}
        \centering
        \includegraphics[width=1\textwidth]{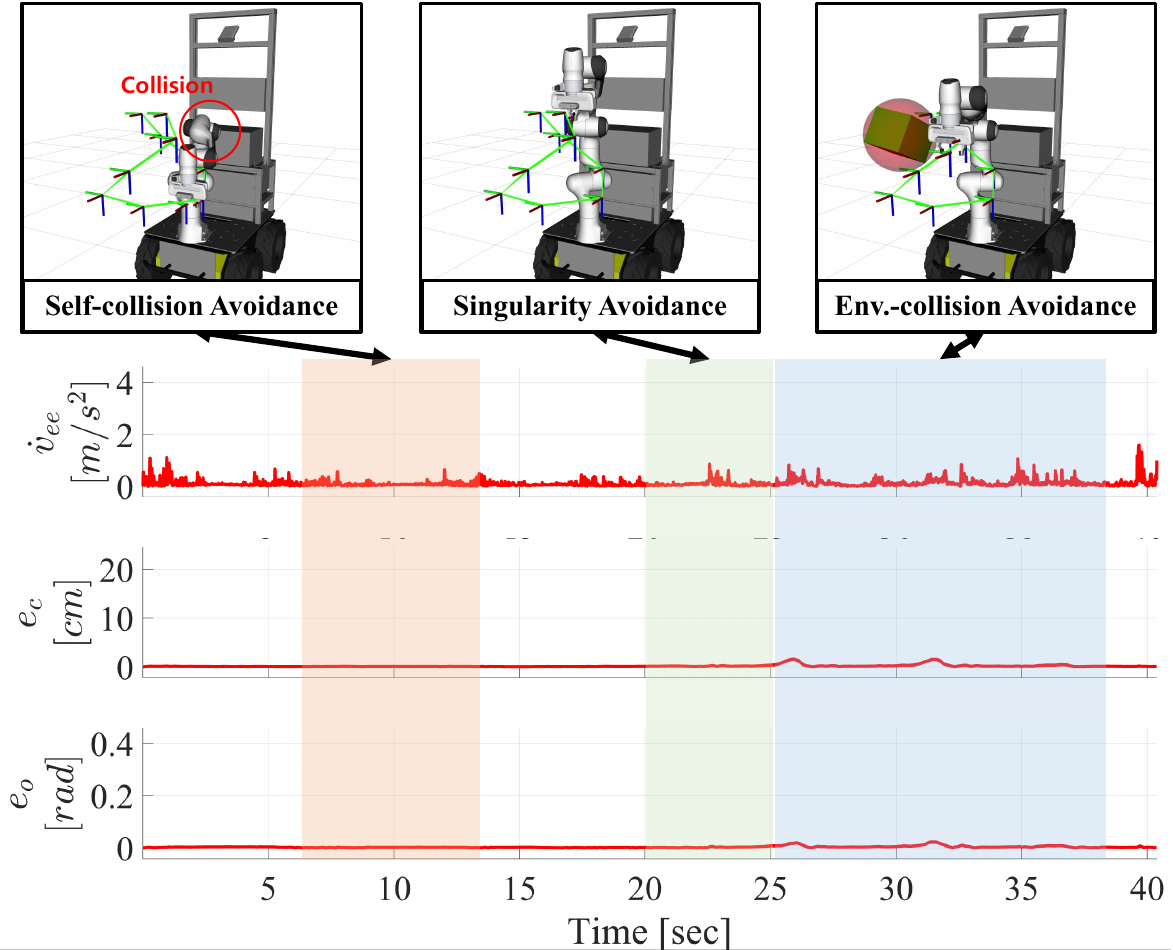}
        \caption{Result of \rmpcc}
        \label{fig:performance_rmpcc}
    \end{subfigure}
    \caption{Comparative demonstration of (a) TT-MPC and (b) \rmpcc under three avoidance scenarios: self-collision, singularity, and environment collision. The top row of images illustrates configuration of the manipulator for each scenario, while the bottom plots show end-effector acceleration (upper), contouring error (bottom), and orientation error (lower). The color shaded in the plots corresponds to self-collision (red), singularity (green), and environment collision avoidance (blue).}
    \label{fig:performance_comparison}
\end{figure*}

\begin{table*}[t!]
    \renewcommand{\arraystretch}{1.0}
    \centering
    \caption{Performance comparison between TT-MPC and RMPCC controllers under self-collision, singularity, and environmental collision scenarios. The table presents the maximum and mean values of the end-effector acceleration $\dot{v}_{ee}$ in m/s$^2$, contouring error $\be_c$ in cm, and orientation error $\be_o$ in rad.}
    \label{tab:performance comparison}
    \begin{tabular}{l l | l l | l l | l l}
    \hline
    \textbf{Problem}  & \textbf{Controller} 
                     & max($\dot{v}_{ee}$) & mean($\dot{v}_{ee}$) 
                     & max($e_c$) & mean($e_c$) 
                     & max($e_o$) & mean($e_o$) \\
    & & [m/s$^2$] & [m/s$^2$] & [cm] & [cm] & [rad] & [rad] \\
    \hline
    Self Collision 
      & TT-MPC & 2.83  & 0.423 & 5.12  & 3.28    & 0.124  & 0.0368 \\
      & \rmpcc & \textbf{0.644} & \textbf{0.0878} & \textbf{0.105} 
               & \textbf{0.0672} & \textbf{2.38E-3} & \textbf{1.254E-3} \\
    \hline
    Singularity    
      & TT-MPC & \textbf{0.450} & \textbf{0.0906} & 22.27 & 8.18 & 0.0108 & \textbf{1.35E-3} \\
      & \rmpcc & 0.865 & 0.0937 & \textbf{0.370} 
               & \textbf{0.158} & \textbf{7.59E-3} & 2.06E-3 \\
    \hline
    Env. Collision 
      & TT-MPC & 4.18 & 0.760 & 16.2 & 9.02 & 0.416 & 0.138 \\
      & \rmpcc & \textbf{1.06} & \textbf{0.138} & \textbf{1.61} 
               & \textbf{0.348} & \textbf{0.0226} & \textbf{5.72E-3} \\
    \hline
    
    \end{tabular}
\end{table*}

\textcolor{black}{The RMPCC is implemented along the reference path shown in Fig. \ref{fig:hardware setup}, with a 16 cm-radius sphere serving as the obstacle. ArUco markers \cite{garrido2014automatic} are attached to the sphere to enable the perception.} In this work, object detection is assumed to be completed independently and prior to the application of the proposed method. In addition, it is assumed that $\bp_{\text{obs}}$ and $r_{\text{obs}}$ accurately represent the true obstacle position and its radius.

\subsection{Baseline and Evaluation Metrics}
The \rmpcc framework was compared with a time parameter based \ttmpc, under identical CBF-based constraints. Other path-parameterized methods such as BoundMPC \cite{oelerich2024boundmpc}, rely on different system models, distinct formulations, and different control frequencies, which makes a direct performance comparison with \rmpcc difficult.

The performance was evaluated using six metrics: motion smoothness (end-effector acceleration $\dot{v}_{ee}$), path-following accuracy (contouring error $e_c = \| \be_c \|$, and orientation error, $e_o = \| \be_o \|$), and safety (manipulability $\mu$, the self-minimum distance, $d_{\text{self}}$, and the environment-minimum distance, $d_{\text{env}}=\min_{l}{d_{\text{env},l}}$).

\subsection{Results and Discussion}
Table~\ref{tab: computation time} shows the computational time for key steps in the \rmpcc framework. The average runtime was 8.71 ms, within the 100 Hz control cycle. Although occasional computation time overruns (exceeding 10 ms) were observed, they occurred only 2–3 times per experiment and were safely handled by reusing the previous control input. Over half of the time was spent computing $d_{\text{env}}$ and its gradient due to the large \nn model.

\begin{table}[h!]
    \renewcommand{\arraystretch}{1.0}
    \centering
    \caption{Statistics of computation time $T_{\text{comp}}$ in ms for each steps.}
    \label{tab: computation time}
    \begin{tabular}{c | c | c | c}
    \hline
      & min($T_{\text{comp}}$) & max($T_{\text{comp}}$)  & mean($T_{\text{comp}}$) \\
    \hline
    Total                                               & 5.96 & 16.7 & 8.71  \\
    Getting $d_{\text{self}}, \nabla_{\bq} d_{\text{self}}$ & 0.56 & 5.05 & 0.72  \\
    Getting $d_{\text{env}}, \nabla_{\bq} d_{\text{env}}$   & 3.71 & 13.6 & 4.90  \\
    Linearization                                       & 0.30 & 4.49 & 0.47  \\
    Solving \ocp                                          & 0.78 & 5.73 & 1.82  \\
    \hline
    \end{tabular}
\end{table}

\begin{figure}[t!]
    \centering
    \includegraphics[width=0.48\textwidth]{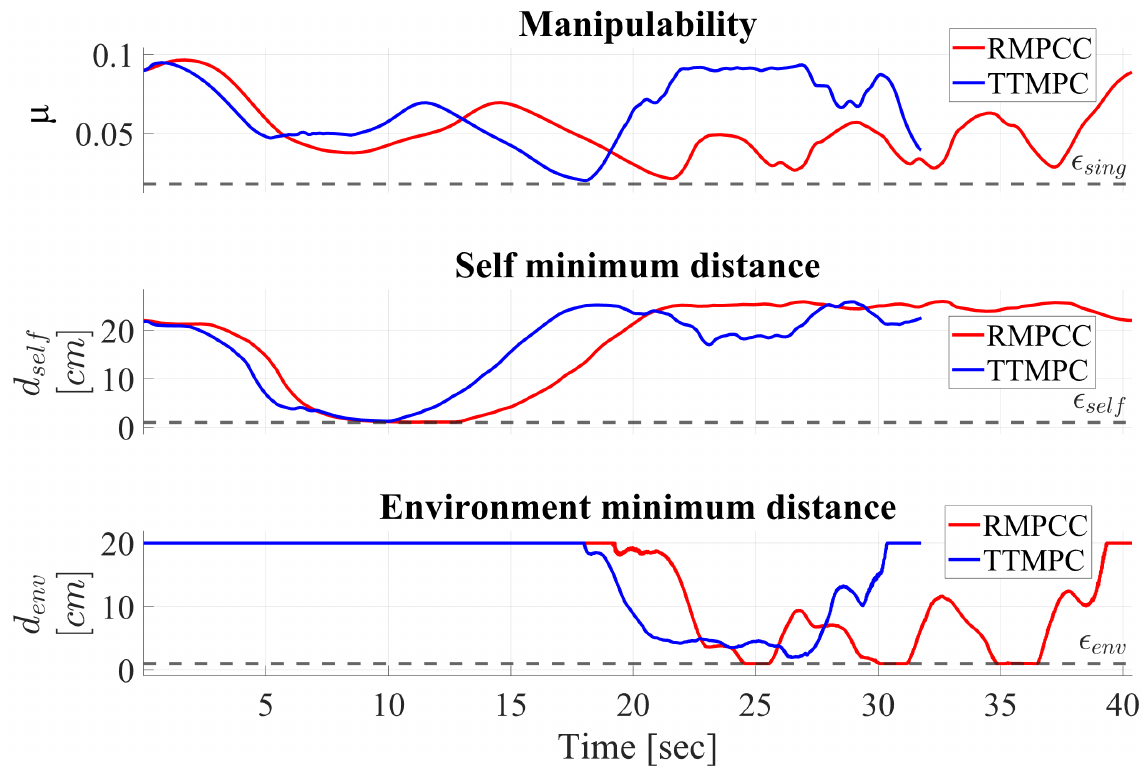}
    \caption{Measured safety metrics during task execution: (i) manipulability index $\mu$, (ii) minimum self-collision distance $d_{\text{self}}$, and (iii) minimum environment distance $d_{\text{env}}$. These remain above the prescribed thresholds $\epsilon_{sing}$, $\epsilon_{\text{self}}$ and $\epsilon_{\text{env}}$, indicating effective singularity and collision avoidance.}
    \label{fig:safety matrices}
\end{figure}

\begin{figure}[t!h!!!]
    \centering
    \includegraphics[width=0.48\textwidth]{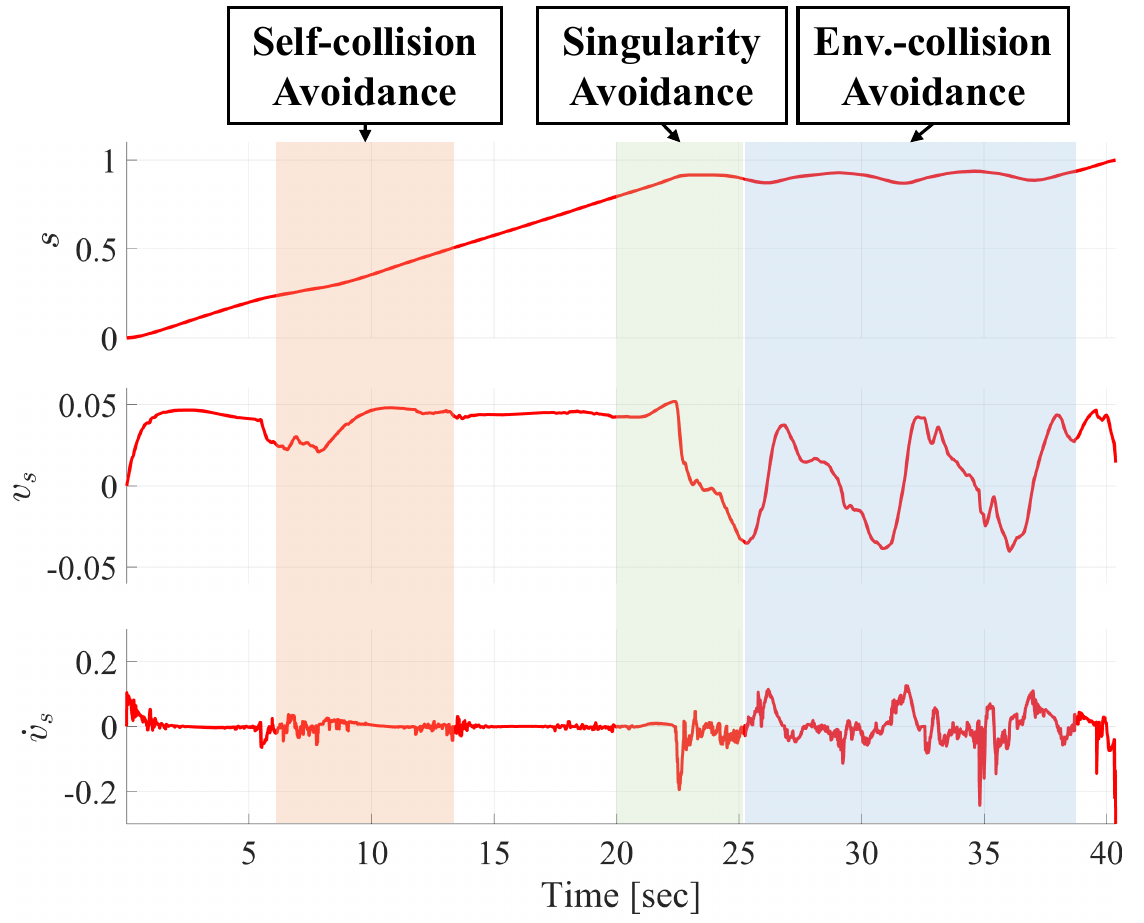}
    \caption{Time evolution of the path parameter $s$, its velocity $v_s$ , and its acceleration $\dot{v}_s$. Each shaded region  denotes a different avoidance scenario: self-collision (red), singularity (green), and environment-collision (blue).}
    \label{fig:path parameter}
\end{figure}

Fig. \ref{fig:safety matrices} shows the evolution of $\mu$, $d_{\text{self}}$, and $d_{\text{env}}$. As the manipulator followed the reference path, three constraints emerged. Between 5$\sim$15 s, $d_{\text{self}}$ decreased as the robot moved backward, bringing links 4 and 5 close to the mobile base (Fig. \ref{fig:performance_comparison}), risking self-collision. Next, the manipulator neared a singular configuration with $\mu$ approaching zero - occurring at 15$\sim$20 s for \ttmpc and 20$\sim$25 s for \rmpcc. Finally, a dynamic external obstacle appeared along the path between 20$\sim$30 s for \ttmpc and 25$\sim$37 s for \rmpcc, creating an environment collision scenario.

Both \ttmpc and \rmpcc enforced safety through \cbfs-based constraints, yet their tracking performance varied. As shown in Table~\ref{tab:performance comparison}, \ttmpc experienced higher end-effector accelerations (up to 4.18 m/s$^2$) with abrupt movements, whereas \rmpcc remained under 1.06 m/s$^2$. Additionally, the maximum contouring and orientation errors for \ttmpc were 22.27 cm and 0.416 rad, compared to only 1.61 cm and 0.0226 rad for \rmpcc.

The real-time adjustment of $s$ in \rmpcc contributes to its improved performance. Since \ttmpc follows a fixed reference trajectory, it cannot easily slow down or reverse to avoid collisions, resulting in larger deviations from the reference and increased tracking errors. In contrast, \rmpcc dynamically adapts 
$s$ based on the condition of the robot (see Fig. \ref{fig:path parameter}), yielding smoother movements and much lower tracking errors.

\section{Conclusion and Outlook} \label{sec:conclusion}
This paper presents the robot manipulator path-following framework \emph{\rmpcc} for dynamic environments. In contrast to existing approaches that consider only the position error and a limited set of safety constraints, the developed controller minimizes both position and orientation errors while avoiding singularities, self-collisions, and external dynamic obstacles. A Gauss-Newton Hessian approximation combined with a Jacobian-based linearization enables reliable operation at 100 Hz. As a result, \rmpcc achieves nearly 10× faster computation than SOTA frameworks, enabling the robot to handle constraints and disturbances in real time.

The validation experiments conducted under real-world conditions demonstrate that low contouring and orientation errors are achieved using the proposed framework. Compared to the baseline experiments performed with the \ttmpc, our \rmpcc method significantly reduces abrupt motion compared to the baseline. These results suggest that \rmpcc is well suited for industrial or collaborative environments demanding safe and precise robot manipulation under dynamic and uncertain operational conditions. 

One limitation of the current framework is its reliance on accurate obstacle perception. Future work will incorporate perception uncertainties (e.g., sensor noise, resolution) directly into the \rmpcc pipeline.

\appendices

\section{Neural Network Architecture for Self Distance Predictor} \label{app.A}
A dataset was generated using the precise geometry of the \textit{Franka Emika Panda} robot (Fig.~\ref{fig:hardware setup}). Uniform random sampling within joint limits produced 10 million configurations: 40\% in collision, 10\% within 5 cm, and 50\% in free space - with signed distances computed by FCL. A fully connected \nn with ReLU activations minimizes RMSE, with increased penalty on near-collision samples while implicitly learning forward kinematics through the feature vector $\bq_{\text{in}} = [\bq, \cos(\bq), \sin(\bq)] \in \R^{3n}$ as input.
The architecture has two hidden layers with 256 and 64 neurons. After 1,000 epochs, the validation RMSE was 3.09 cm overall (1.66 cm in free space, 1.24 cm near-collision, and 4.5 cm in collision cases).

\section{Derivation of Jacobian of Orientation error} \label{app.B}
The first-order approximation of $\be_o$ at $s_{\text{init}}$ and $\bq_{\text{init}}$ \textcolor{black}{is derived} in order to compute the Jacobian. 
\begin{equation}
\begin{aligned}
    &\be_o(\bx) \stackrel{  \eqref{L_ori}}{=} \Log(\bR_{r}(s)^\top \bR_{ee}(\bq) ) \\
    & \stackrel{(a)}{=} \Log\bigl[\Exp(\delta \phi)\,\bar{\bR}_{r}\bigr]^\top
    \,\bigl[\Exp(\delta \varphi)\,\bar{\bR}_{ee}\bigr] \\
    & \stackrel{(b)}{=} \Log\Bigl[\bar{\bR}_{r}^\top \Exp\bigl(-\bar{\bm{\phi}^{\prime}_{i}}\,\delta s \bigr)^\top   
    \Exp\bigl(\bar{\bJ}_{\text{ori}}\,\delta \bq\bigr) \bar{\bR}_{ee}  \Bigr] \\
    & \stackrel{  \eqref{exp property}}{=} \Log\Bigl[\bar{\bR}_{r}^\top \bar{\bR}_{ee} \,
    \Exp\bigl(-\bar{\bR}_{ee}^\top \bar{\bm{\phi}^{\prime}_{i}}\,\delta s\bigr)
    \Exp\bigl(\bar{\bR}_{ee}^\top \bar{\bJ}_{\text{ori}}\,\delta \bq\bigr)\Bigr] \\
    & \stackrel{(c)}{\approx} \bar{\be}_o
    - \bJ_r^{-1}\!\bigl(\bar{\be}_o\bigr)\,\bar{\bR}_{ee}^\top \bar{\bm{\phi}^{\prime}_{i}}\,\delta s
    + \bJ_r^{-1}\!\bigl(\bar{\be}_o\bigr)\,\bar{\bR}_{ee}^\top \bar{\bJ}_{\text{ori}}\,\delta \bq
\end{aligned}
\end{equation}

In step~(a), $\bR_{r}(s)$ and $\bR_{ee}(\bq)$ follow infinitesimal rotations $\delta \phi,\delta \varphi \in \R^3$ from their initial values $\bar{\bR}_{r} = \bR_{r}(s_{\text{init}})$ and $\bar{\bR}_{ee} = \bR_{ee}(\bq_{\text{init}})$ (see \eqref{Exp R}), due to changes $\delta s,\delta \bq$.  
In step~(b), $\bar{\bm{\phi}_{i}^{\prime}}$ is defined as $\bigl.\tfrac{d\bm{\phi}_i}{ds}\bigr|_{s=s_{\text{init}}}$ with $s_i \le s_{\text{init}} \le s_{i+1}$ from \eqref{orientation spline}. Also the skew-symmetric transpose $\bigl(\mathbf{a}^\wedge\bigr)^\top = -\mathbf{a}^\wedge$ and $\bigl(\exp\mathbf{A}\bigr)^\top = \exp(\mathbf{A}^\top)$ are applied.  
Finally, step~(c) uses the first-order approximation of $\Log(\cdot)$ \eqref{Exp Log Jacobian} for $-\bar{\bR}_{ee}^\top \bar{\bm{\phi}_{i}^{\prime}}\delta s$ and $\bar{\bR}_{ee}^\top \bar{\bJ}_{\text{ori}}\delta \bq$



\bibliographystyle{IEEEtran}
\bibliography{references}

\end{document}